\documentclass{article}

\usepackage{PRIMEarxiv}

\usepackage[utf8]{inputenc} % allow utf-8 input
\usepackage[T1]{fontenc}    % use 8-bit T1 fonts
\usepackage{hyperref}       % hyperlinks
\usepackage{url}            % simple URL typesetting
\usepackage{booktabs}       % professional-quality tables
\usepackage{amsfonts}       % blackboard math symbols
\usepackage{nicefrac}       % compact symbols for 1/2, etc.
\usepackage{microtype}      % microtypography
\usepackage{fancyhdr}       % header
\usepackage{graphicx}       % graphics
\usepackage{amsmath}        % used to define argmax commands
\usepackage{amssymb}        % used to get triangle equality sign
\usepackage{caption}        % table captions
\usepackage{floatrow}   
\usepackage{subcaption}
\usepackage[framemethod=tikz]{mdframed}
\graphicspath{{media/}}     % organize your images and other figures under media/ folder
\DeclareMathOperator*{\argmax}{argmax}
\DeclareMathOperator*{\argmin}{argmin}

\title{A1 SLAM: Quadruped SLAM using the A1's onboard sensors}

\author{
  Jerred Chen \\
  Georgia Institute of Technology \\
  Atlanta, GA \\
  \texttt{jchen788@gatech.edu} \\
  \And
  Frank Dellaert \\
  Georgia Institute of Technology \\
  Atlanta, GA \\
  \texttt{frank.dellaert@cc.gatech.edu}
}

\begin{document}
\maketitle

\begin{abstract}
Quadrupeds are robots that have been of interest in the past few years due to their versatility in navigating across various terrain and utility in several applications. For quadrupeds to navigate without a predefined map a priori, they must rely on SLAM approaches to localize and build the map of the environment. Despite the surge of interest and research development in SLAM and quadrupeds, there still has yet to be an open-source package that capitalizes on the onboard sensors of an affordable quadruped. This motivates the A1 SLAM package, which is an open-source ROS package that provides the Unitree A1 quadruped with real-time, high performing SLAM capabilities using the default sensors shipped with the robot. A1 SLAM solves the PoseSLAM problem using the factor graph paradigm to optimize for the poses throughout the trajectory. A major design feature of the algorithm is using a sliding window of fully connected LiDAR odometry factors. A1 SLAM has been benchmarked against Google's Cartographer and has showed superior performance especially with trajectories experiencing aggressive motion.
\end{abstract}

% keywords can be removed
\keywords{Robotics \and SLAM \and Quadruped}

\begin{figure}[h]
\centering
\includegraphics[width=9cm]{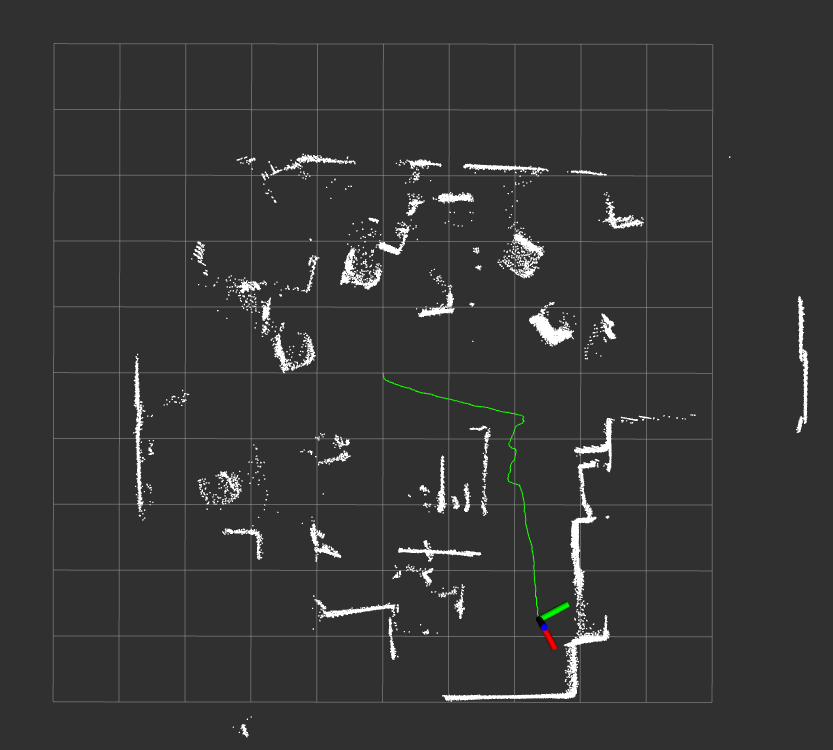}
\captionsetup{justification=centering}
\caption{The resulting map produced from the A1 SLAM algorithm.}
\end{figure}

\section{Introduction}

Quadrupeds are highly versatile robots that can traverse over difficult terrain that wheeled mobile robots are unable to. This flexibility makes quadrupeds appealing for various applications, such as inspection, surveying construction sites, and search-and-rescue. However, to effectively perform these tasks autonomously, quadrupeds, as with other mobile robots, require a form of perception that will enable them to localize when placed in an environment without a priori knowledge. For robots to know its location in the environment, it must localize against a predefined map, but a robot can only create a map based on its known location. To solve this chicken-and-egg problem, simultaneous localization and mapping, or SLAM, is the standard approach used for mobile robots by optimizing for the robot's location and map simultaneously. The estimated poses and map from SLAM algorithms can then be used for downstream tasks such as facilitating controllers depending on the terrain or planning in navigation. Despite the recent developments in both quadruped robotics and in SLAM research, there has yet to be an open-source package that is specifically designed for high performing SLAM on quadrupeds. The contribution of this work is twofold:
\begin{itemize}
\item A real-time 2D SLAM algorithm that outperforms some of the most commonly used out-of-the-box SLAM algorithms publicly available
\item An open-source ROS package which aims for convenient and simple use with the onboard sensors on any A1 quadruped, available at \url{https://github.com/jerredchen/A1_SLAM}
\end{itemize}

\section{Background and Related Work}

\subsection{SLAM and Factor Graphs}

Although SLAM has been an ongoing area of research for nearly two decades, the fundamental problem of recovering pose values that maximized the posterior probability remained the same. Initial iterations of SLAM consisted of filtering approaches, such as Extended Kalman Filters or Rao-Blackwellised Filters \cite{durrant2006simultaneous}, to reduce computation for real-time capabilities. An alternative to filtering was smoothing, which could be achieved with using factor graphs \cite{dellaert2017factor}. Factor graphs efficiently estimated values across entire trajectories by exploiting sparsity in the SLAM problem. Using factor graphs for real-time SLAM became prevalent due to advances in efficient, incremental inference with Square Root SAM \cite{dellaert2006square}, iSAM \cite{kaess2008isam}, and iSAM2 \cite{kaess2012isam2}. Factor graph SLAM solutions are now commonly among the state-of-the-art, with open-sourced packages such as LIO-SAM \cite{liosam2020shan}.

\subsection{Quadruped SLAM}

Recent years have shown a surge of interest in extending SLAM capabilities to quadrupeds due to their versatility to navigate across various terrains. \cite{ramezani2020online} showed how quadrupeds could navigate in adverse environment conditions with deep learned loop closure. \cite{wisth2022vilens} demonstrate how vision, LiDAR, IMU, and legged odometry could be incorporated into a single factor graph framework for robust SLAM solution. Robust LiDAR mapping for long durations during the DARPA SubTerranean Challenge on a quadruped is exhibited in \cite{reinke2022iros}.

The motivation for this package is not to develop a novel SLAM algorithm which beats the state-of-the-art; rather, the A1 SLAM package was released to be a high quality and robust open-source SLAM solution designed for the Unitree A1 quadruped using the default onboard sensors.

\section{Factor Graph Optimization for PoseSLAM}

\begin{figure}[htbp]
\centering
\includegraphics[width=7.5cm]{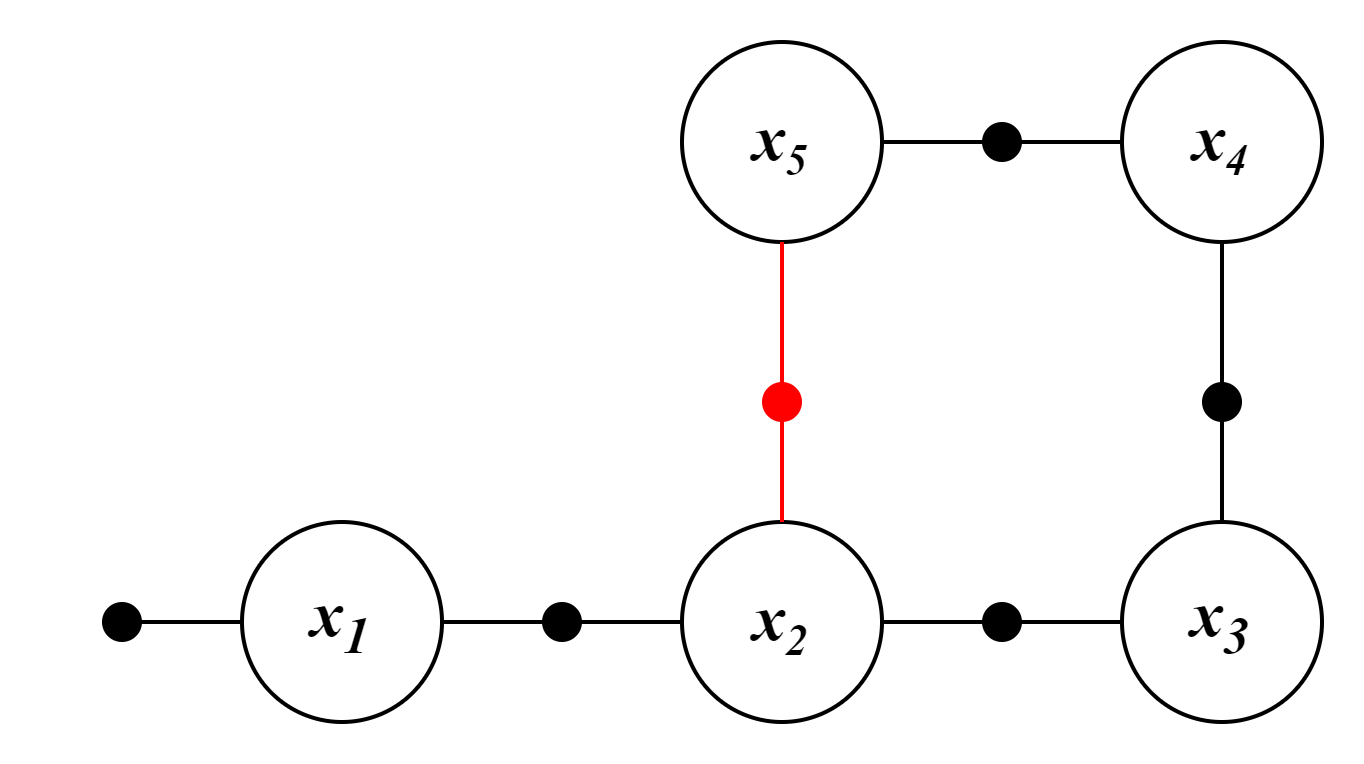}
\captionsetup{justification=centering}
\caption{An example of a PoseSLAM factor graph. Note that the red factor represents a loop closure, which enforces a constraint between $x_5$ and $x_2$.}
\label{fig:pose_slam}
\end{figure}

The A1 SLAM algorithm employs a factor graph to solve the PoseSLAM problem - explicitly optimizing the poses of the robot first and constructing the map afterwards. A factor graph is a bipartite graph which consists of variables $x_i$, states to be optimized, and factors $\phi_i$, probabilistic constraints on $X_i$ where $X_i \subseteq \{x_1, x_2, \dots, x_m\}$. Pictured in Figure \ref{fig:pose_slam} is an example of a factor graph depiction for a PoseSLAM problem. Ultimately, the goal of PoseSLAM is to obtain $X^*$, the poses that maximizes the posterior probability of $X$ given noisy sensor measurements. In the factor graph framework, the global function
$$\phi(X) = \phi_1(X_1)\phi_2(X_2) \dots \phi_n(X_n) = \prod_i^n \phi_i(X_i)$$
represents the unnormalized posterior distribution, with the factor graph depicting the factorization of $\phi(X)$. To recover $X^*$, we perform maximum a posteriori (MAP) inference, which can be formulated as:
$$ X^* = \argmax_X \phi(X) = \argmax_X \prod_i \phi_i(X_i) $$
If we assume that all sensor measurements are corrupted with zero-mean Gaussian noise, we obtain a nonlinear least squares optimization problem:
$$ X^* = \argmin_X \sum_i \|h(X_i) - z_i\|^2_{\Sigma_i}$$
where $h$ is the sensor model likelihood, and $z_i$ and $\Sigma_i$ are the measurement and covariance associated with $X_i$, respectively. Since we are optimizing the 2D poses of the robot $x$ where $x$ lies on the $SE(2)$ manifold, standard nonlinear optimization approaches cannot trivially optimize for $X$ without leaving the manifold. Because manifolds are locally homeomorphic to Euclidean space, i.e. for each point on $SE(2)$ there exists a local neighborhood where the exponential map $\mathbb{R}^3 \rightarrow SE(2)$ (and logarithmic map $SE(2) \rightarrow \mathbb{R}^3$) exists, we instead optimize for local coordinates $\xi \in \mathbb{R}^3$, where $\xi \doteq (\delta x, \delta y, \delta\theta)$. Linearizing the likelihood function and minimizing the objective function with respect to $\xi$ results in:
$$\Xi^* = \argmin_\Xi \|h(X_i) + H_i\xi_i - z_i\|^2_{\Sigma_i}$$
where $H_i$ is the Jacobian of the likelihood function and $\Xi \triangleq \{\xi_i\}$. More information about optimizing on manifolds can be found in \cite{dellaert2017factor}.

\section{A1 SLAM Implementation}

\begin{figure}[h]
\centering
\includegraphics[width=12.5cm]{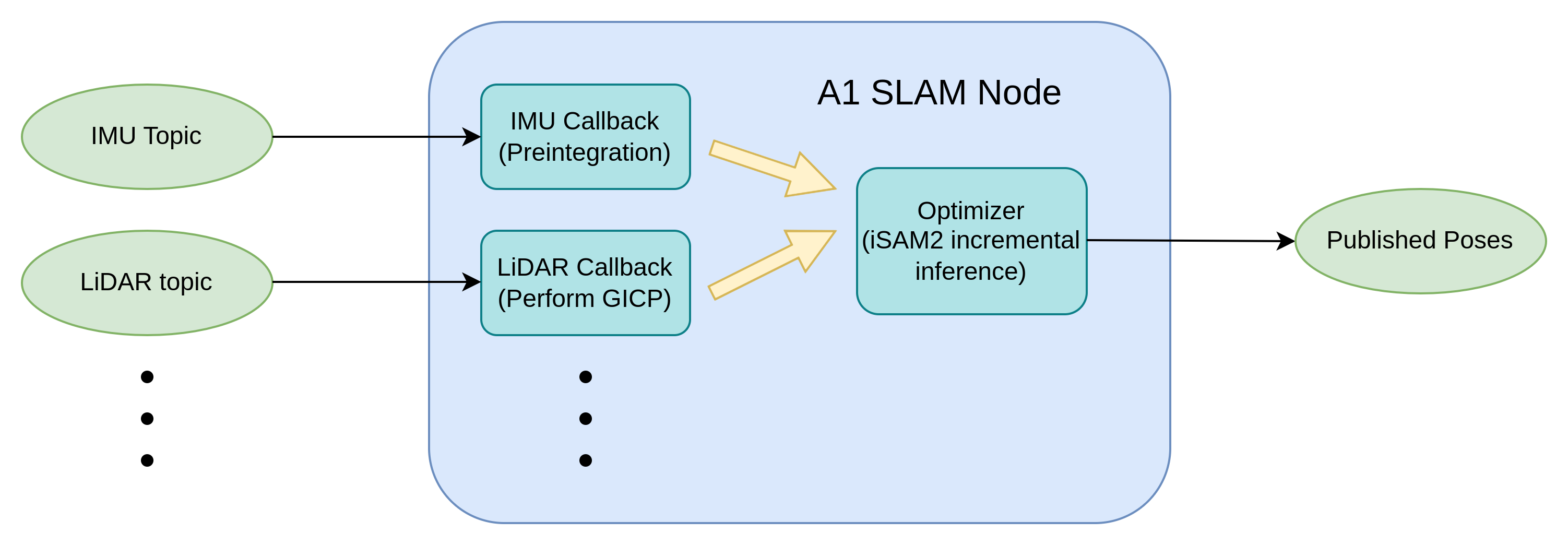}
\captionsetup{justification=centering}
\caption{The software implementation of the A1 SLAM algorithm.}
\label{fig:software}
\end{figure}

Figure \ref{fig:software} shows the software implementation of the A1 SLAM algorithm. Each sensor callback subscribes to their respective sensor topic which receives the incoming measurements. The callback will perform any necessary data preprocessing and generate factors that will be added to a global pose graph optimizer. The optimizer uses iSAM2 to perform incremental inference with the incoming factors. Each iSAM2 update is followed by the current pose estimate of the robot being published, with all past poses in the trajectory optionally published at a lower frequency. The following section specifies (ongoing) implementation details for each sensor callback, which are then fused into a single SLAM optimization problem. To implement all factor graph data structures and optimizations in the A1 SLAM algorithm, the GTSAM library \cite{gtsam} was used.

\begin{figure}[h]
    \centering
    \begin{minipage}{0.5\textwidth}
        \centering
        \includegraphics[width=8cm]{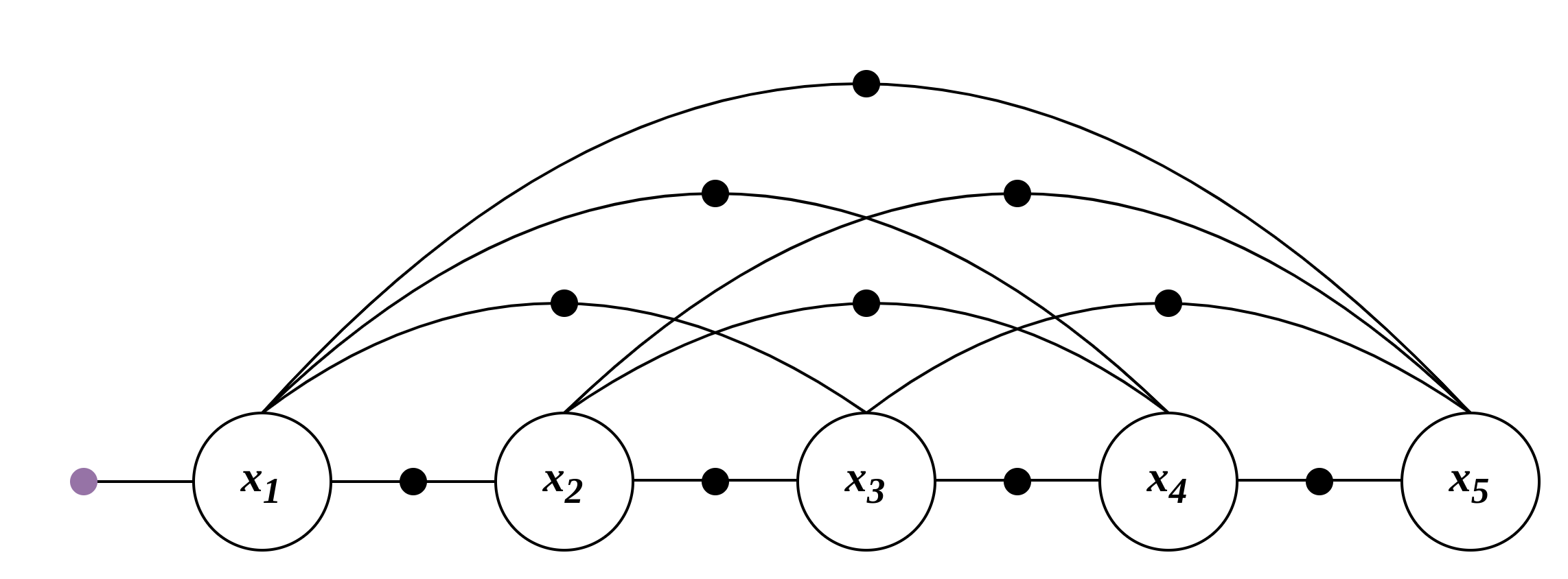}
    \end{minipage}

\captionsetup{justification=centering}
\caption{An example of a factor graph for 5 poses with fully connected skip factors.}
\label{fig:factor_graph}
\end{figure}

\subsection{2D LiDAR}

The main design feature of the A1 SLAM algorithm is incorporating a sliding-window of fully connected skip factors. The A1 quadruped comes with a SLAMTEC RPLIDAR sensor\footnote{\url{https://www.slamtec.com/en/Lidar/Mapper}} that allows for 2D mapping capabilities. Some traditional approaches use a scan-to-scan method such as ICP to estimate the odometry between two consecutive LiDAR scans, which results in two major drawbacks: scan-to-scan approaches tend to accumulate error quickly, and simply estimating odometry between consecutive scans did not capitalize on the factor graph paradigm. As seen in Figure \ref{fig:factor_graph}, skip factor connections are added to further constrain the poses of the robot over a set number of poses, which serve as an implicit loop closure. The sliding window of fully connected skip factors mitigates the rapid accumulation of error. To compute these binary pose factors, the off-the-shelf multithreaded GICP implementation Fast-GICP \cite{koide2021voxelized} is used to estimate the odometry of the robot given two LiDAR scans.

\section{Experimental Results}

\subsection{Experiment Setup}

The A1 quadruped was remote controlled to walk a trajectory within a roughly 5 meter by 5 meter area. Four trajectories with varying translational and rotational velocities were perform, where sensor measurements including 2D LiDAR ranges, IMU data, and depth images were recorded in a rosbag. Motion capture cameras were also used to obtain a ground truth trajectory for each run. Table \ref{tab:traj_info_table} provides basic details on each trajectory dataset that was collected. The average velocity was calculated simply from the total distance and duration of the run. Note that trajectories 2 and 3 are marked with more aggressive translation and rotation; the quadruped walked with an average velocity of approximately 2-3 times the velocity in trajectories 1 and 2.

\begin{table}[htbp]
    \centering
    \captionsetup{justification=centering}
    \caption{Information on each trajectory that was recorded.}
    \begin{tabular}{cccc}
        \toprule
        Trajectory & Duration (s) & Total Distance (m) & Average Velocity (m/s) \\
        \midrule
        1 & 34.5 & 5.3 & 0.15 \\
        2 & 14.4 & 5.0 & 0.34 \\
        3 & 22.4 & 5.0 & 0.22 \\
        4 & 44.3 & 5.2 & 0.12 \\
        \bottomrule
    \end{tabular}
    \label{tab:traj_info_table}
\end{table}

The A1 SLAM algorithm was evaluated against Google's out-of-the-box Cartographer \cite{hess2016real}, which was identified in \cite{Yagfarov18icarcv_2dslam_map_comparison}\cite{Zou22its_analysis_lidar_slam}\cite{Fan21jpcs_2dslam_eval} to be the highest performing algorithm out of a conducted survey of 2D LiDAR SLAM approaches. To maintain consistency, only 2D LiDAR measurements were used to estimate the trajectory for both A1 SLAM and Cartographer. For each trajectory, the estimated pose values were outputted from the A1 SLAM and Cartographer algorithms and evaluated against the ground truth poses. The metric used for evaluation was the absolute pose error (APE) metric, where the APE between the SLAM estimated pose $P_{est,i}$ and the reference ground truth pose $P_{ref,i}$ is defined as:
$$APE_i = P^{-1}_{ref,i} P_{est,i} \in SE(3)$$
To do this, each pose trajectory was downsampled to 10 Hz, the frequency of the A1 SLAM algorithm, and then aligned with the ground truth trajectory before measuring the APE between the mocap pose and SLAM estimated pose. The evo \footnote{\url{https://github.com/MichaelGrupp/evo}} package was used to evaluate the APE. 

The speed of both algorithms were also benchmarked. The elapsed wall clock times of offline processing the datasets were compared against each other and the total duration of the dataset, which gives insight in the maximum frequency that these algorithms can operate at. Both A1 SLAM and Cartographer were run on a computer with an Intel Core i7-8550U CPU @ 1.80GHz.

\subsection{Results and Discussion}

\subsubsection{Accuracy Evaluation}

\begin{table}[h]
    \centering
    \captionsetup{justification=centering}
    \caption{RMSE APE evaluated for Cartographer and A1 SLAM trajectories.}
    \begin{tabular}{lcccc}
        \toprule
        Algorithm & Trajectory 1 & Trajectory 2 & Trajectory 3 & Trajectory 4 \\
        \midrule
        Cartographer & \textbf{0.040} & 0.962 & 1.397 & 0.134 \\
        A1 SLAM (ours) & 0.047 & \textbf{0.118} & \textbf{0.145} & \textbf{0.043} \\
        \bottomrule
    \end{tabular}
    \label{tab:ape_table}
\end{table}

\begin{figure}[htbp]
\centering
\begin{minipage}{0.35\textwidth}
    \centering
    \includegraphics[width=5cm]{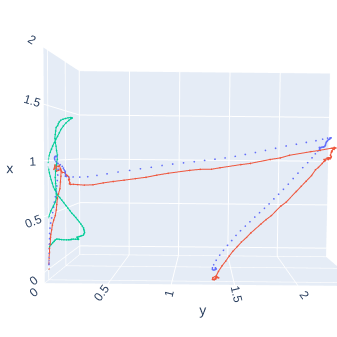}
\end{minipage}
\begin{minipage}{0.3\textwidth}
    \centering
    \includegraphics[width=4cm]{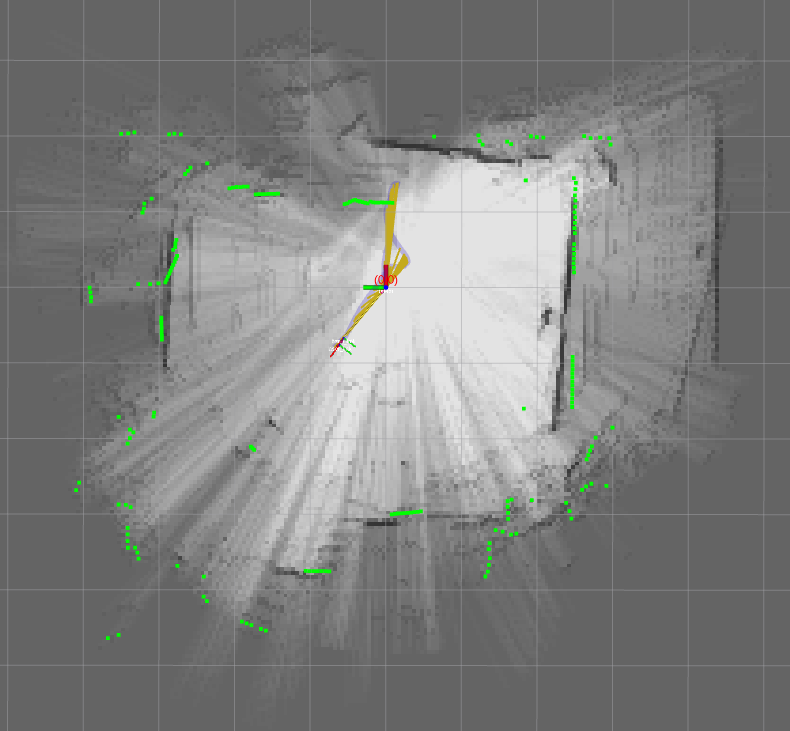}
\end{minipage}
\begin{minipage}{0.3\textwidth}
    \centering
    \includegraphics[width=4cm]{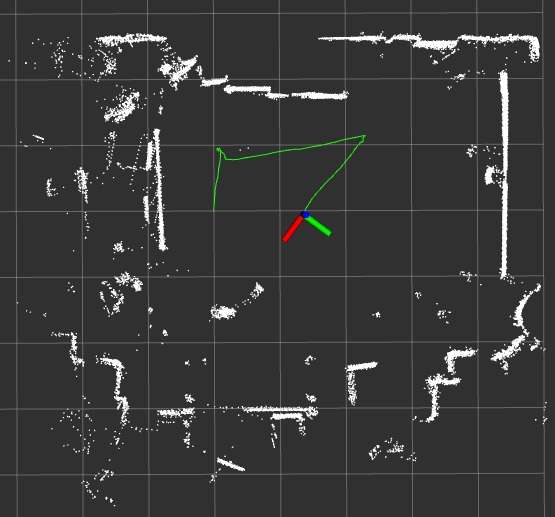}
\end{minipage}
\bigskip
\begin{minipage}{0.35\textwidth}
    \centering
    \includegraphics[width=5cm]{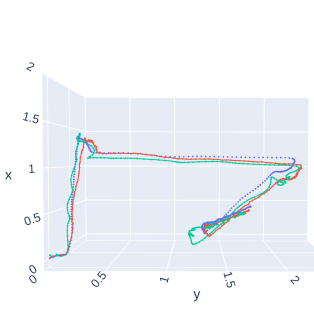}
\end{minipage}
\begin{minipage}{0.3\textwidth}
    \centering
    \includegraphics[width=4cm]{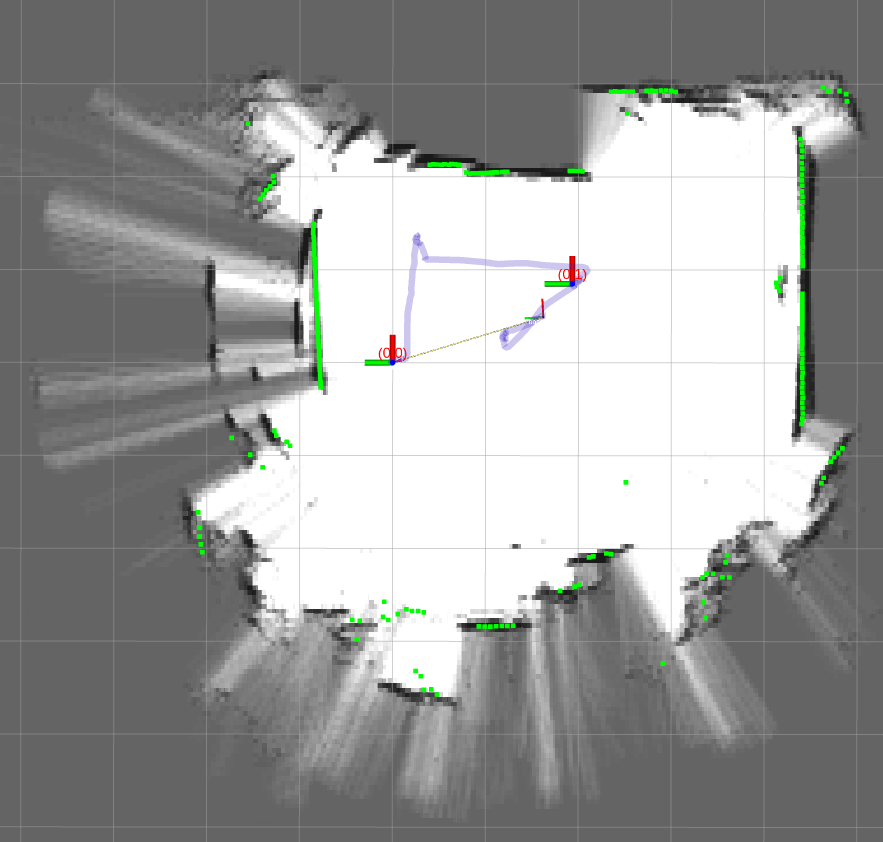}
\end{minipage}
\begin{minipage}{0.3\textwidth}
    \centering
    \includegraphics[width=4cm]{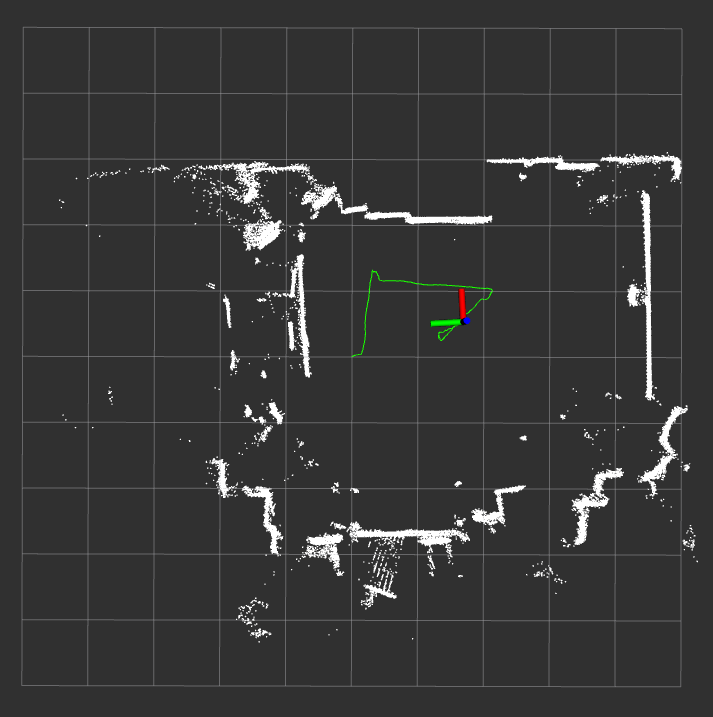}
\end{minipage}
\captionsetup{justification=centering}
\caption{The resulting trajectories and map that were produced for the third experiment (top row) and fourth experiment (bottom row). Red trajectory is A1 SLAM, green is Cartographer, blue is ground truth. Middle column is Cartographer produced map, right column is A1 SLAM produced map.}
\label{fig:trajectories}
\end{figure}

Table \ref{tab:ape_table} details the RMSE (root-mean-square error) APE for the A1 SLAM and Cartographer trajectories. In the first trajectory, Cartographer and A1 SLAM provide nearly identical results, with Cartographer yielding slightly less error than the A1 SLAM. In the remaining experiments, A1 SLAM vastly outperformed Cartographer, particularly for the trajectories that had noticeably more aggressive translation and/or rotation. 

Figure \ref{fig:trajectories} shows the trajectories and map that were produced for the third and fourth experiment. In the fourth experiment (shown in the bottom row), A1 SLAM trajectory qualitatively aligned better with the mocap trajectory, which is reflected since A1 SLAM had significantly less RMSE APE than Cartographer. The Cartographer produced is seen to have wall misalignment artifacts that appear, particularly in the top right corner, whereas the A1 SLAM map produced well-defined walls throughout the map. The third experiment (shown in the top row) involved aggressive translation, which Cartographer failed to handle. In contrast, A1 SLAM is able to robustly produce a trajectory with RMSE APE of 0.145. The map similarly reflects the poor localization performance from Cartographer and the well-produced map from A1 SLAM.

It is important to note that the evaluation was performed against an out-of-the-box Cartographer algorithm, where additional tuning may be able to result in better performance. However, because of the heavy complexity in the algorithm, tuning Cartographer parameters is not immediately obvious and can be a tedious process. In contrast, A1 SLAM only requires minimal parameter tuning and is designed to function as a black-box algorithm for the A1 quadruped.

\subsubsection{Speed Evaluation}

Although the A1 SLAM algorithm yields significantly less error than Cartographer, Cartographer outperforms A1 SLAM in terms of speed. This can be seen in Table \ref{tab:wall_time_table}, which shows that Cartographer performs its offline SLAM at approximately 10 times faster than A1 SLAM does. This is due to the design of the A1 SLAM algorithm. While Cartographer is written in C++, A1 SLAM is written in Python, which does not experience any compiler optimizations and suffers from extremely slow branching logic. Since A1 SLAM requires many fully connected pose constraints before optimizing a new pose, this requires a Python `for' loop for every new pose which drastically slows down performance. The elapsed wall time of A1 SLAM increases with the length of the submap (the number of scans that experience fully connected skip factors), resulting in a trade-off between speed and accuracy. In addition, A1 SLAM operates effectively single-threaded. Although A1 SLAM utilizes a separate callback for processing LiDAR scans and optimizing the factor graph, the Python GIL forces the algorithm to run effectively on a single-thread. Cartographer, in contrast, uses several background threads to perform global optimization and large loop closures. Despite not performing at the same speed as Cartographer, Table \ref{tab:wall_time_table} reflects that A1 SLAM still performs several times faster than real-time. A1 SLAM only publishes poses at the same frequency as the LiDAR (10 Hz), but it can be seen to operate at frequencies greater than 30 Hz.

Table \ref{tab:optimize_table} shows the underlying optimizer call times for each LiDAR scan. While A1 SLAM's optimization is done through GTSAM, Cartographer performs all of its optimizations using Ceres. In A1 SLAM, necessary relative pose factors are added to the factor graph, an iSAM2 update is made, and all poses in the trajectory are subsequently estimated for each LiDAR scan. Ceres, on the other hand, optimizes the scan-to-map pose given a received LiDAR scan. Table \ref{tab:optimize_table} shows that the underlying optimization calls are both approximately 1 ms, demonstrating how neither optimizations serve as a bottleneck for real time performance for a LiDAR sensor running at 10  Hz.

\begin{table}[ht]
    \centering
    \captionsetup{justification=centering}
    \caption{Elapsed Wall Time of Offline SLAM}
    \begin{tabular}{cccc}
        \toprule
        Trajectory & Dataset Duration (s) & Cartographer Wall Time (s) & A1 SLAM Wall Time (s) \\
        \midrule
        1 & 34.5 & \textbf{0.88} & 10.3 \\
        2 & 14.4 & \textbf{0.35} & 3.7 \\
        3 & 22.4 & \textbf{0.48} & 5.7 \\
        4 & 44.3 & \textbf{1.20} & 12.8 \\
        \bottomrule
    \end{tabular}
    \label{tab:wall_time_table}
\end{table}

\begin{table}[ht]
    \centering
    \captionsetup{justification=centering}
    \caption{Average Optimization Call Time per LiDAR Scan}
    \begin{tabular}{ccc}
        \toprule
        Algorithm & Optimizer & Average Call Time (ms) \\
        \midrule
        A1 SLAM & GTSAM & 1.65 \\
        Cartographer & Ceres & 0.96 \\
        \bottomrule
    \end{tabular}
    \label{tab:optimize_table}
\end{table}

\section{Limitations and Future Work}

The A1 SLAM package is meant to be continuously updated and maintained, so it is imperative to acknowledge its several limitations and possible future directions to improve its performance and usability. 

\subsection{2D LiDAR}
There are several sources of improvement that can be made to the 2D LiDAR callback to improve its utility. As mentioned before, the sliding window is not determined by a distance threshold but by a set number of LiDAR scans that were received. This results in unnecessary computation that is performed, especially when the robot is moving very slowly or is stationary. This change in distance can be detected by using an IMU which can then extract important key frames in the LiDAR scans. Because solving the PoseSLAM problem involves transforming the LiDAR scans into the world frame to obtain the map, a map data structure is not produced which cannot be used for navigation purposes. A potential direction will consist of having a occupancy grid that is continuously updated on a separate thread.

\subsection{Additional Sensors}
Multiple sensor modalities can be fused to further refine the SLAM solution. An IMU is on the A1 quadruped, where preintegrated IMU factors as seen in \cite{forster2016manifold} can be fused to improve the odometry estimates. A depth camera is also on the A1, so both visual odometry approaches and depth point cloud registration can be performed. To estimate the odometry using noisy depth point clouds, certifiably robust registration algorithms such as TEASER++ \cite{Yang20tro-teaser} have been experimented with to produce accurate results. In addition, joint encoder measurements are also available on the A1 quadruped in its "low state" mode. This can be used to fuse legged kinematic constraints in the factor graph and improve the performance.

\subsection{Computational Performance}
As discussed previously, the A1 SLAM processing speed greatly lags behind that of Cartographer. The biggest computational bottleneck is due to much of the processing being done in Python, so porting the code from Python to C++ would result in a dramatic increase in speed. The Python GIL also restricts the performance and introduces additional computational overhead. While in C++, the 2D LiDAR callback could be further improved by having a separate thread optimize the submap of fully connected pose constraints.

\section{Conclusion}
We present A1 SLAM, an open-source ROS package designed for high performing and robust SLAM using the onboard sensors of the Unitree A1 quadruped. A PoseSLAM solution is produced by creating and solving a factor graph using the GTSAM library. A major design feature of the A1 SLAM algorithm is the sliding window of fully connected pose factors that uses a fast, multithreaded GICP method to estimate odometry from two LiDAR scans. The algorithm was benchmarked against Google's Cartographer and yields superior performance especially with trajectories that experience aggressive motion. A1 SLAM is an ongoing project which will be continuously maintained and improved to integrate new features and iterate upon for better performance.

%Bibliography
\bibliographystyle{unsrt}  
\bibliography{references}

\end{document}